\documentclass[lettersize,journal]{IEEEtran}
\usepackage{amsmath,amsfonts}
\usepackage{algorithmic}
\usepackage{algorithm}
\usepackage{array}
\usepackage[caption=false,font=normalsize,labelfont=sf,textfont=sf]{subfig}
\usepackage{textcomp}
\usepackage{stfloats}
\usepackage{url}
\usepackage{verbatim}
\usepackage{graphicx}
\usepackage{booktabs}
\usepackage{cite}
\usepackage{multirow} 
\usepackage{subcaption} 
\begin{document}

\title{{D2-Mamba: Dual-Scale Fusion and Dual-Path Scanning with SSMs for Shadow Removal}}

\author{Linhao Li, Boya Jin, Zizhe Wang, Lanqing Guo$^{\star}$, Hao Cheng$^{\star}$, Bo Li, Yongfeng Dong
\thanks{This research work is partially supported by NSFC under Grant No. 62506115 and by the Natural Science Foundation of Tianjin, China, under the grant 24JCQNJC01230.
This research work is also partially supported by Science Research Project of Hebei Education Department under the grant BJ2025001, BJ2025004.
This research work is also partially supported by the Natural Science Foundation of Hebei Province under Grant F2025202020, F2025202039. (Corresponding Authors: Lanqing Guo, Hao Cheng.)

Linhao Li, Boya Jin, Zizhe Wang, Hao Cheng, Yongfeng Dong are with the School of Artificial Intelligence, Hebei University of Technology, China. Lanqing Guo is with the School of Electrical and Computer Engineering, University of Texas at Austin, USA.
Bo Li is with the School of Health Sciences and Biomedical Engineering, Hebei University of Technology, China.
}

}


\markboth{Journal of \LaTeX\ Class Files,~Vol.~14, No.~8, August~2021}%
{Shell \MakeLowercase{\textit{et al.}}: A Sample Article Using IEEEtran.cls for IEEE Journals}


\maketitle

\begin{abstract}
Shadow removal aims to restore images that are partially degraded by shadows, where the degradation is spatially localized and non-uniform. Unlike general restoration tasks that assume global degradation, shadow removal can leverage abundant information from non-shadow regions for guidance. However, the transformation required to correct shadowed areas often differs significantly from that of well-lit regions, making it challenging to apply uniform correction strategies. This necessitates the effective integration of non-local contextual cues and adaptive modeling of region-specific transformations.
To this end, we propose a novel Mamba-based network featuring dual-scale fusion and dual-path scanning to selectively propagate contextual information based on transformation similarity across regions. Specifically, the proposed Dual-Scale Fusion Mamba Block (DFMB) enhances multi-scale feature representation by fusing original features with low-resolution features, effectively reducing boundary artifacts. The Dual-Path Mamba Group (DPMG) captures global features via horizontal scanning and incorporates a mask-aware adaptive scanning strategy, which improves structural continuity and fine-grained region modeling. Experimental results demonstrate that our method significantly outperforms existing state-of-the-art approaches on shadow removal benchmarks.
\end{abstract}

\begin{IEEEkeywords}
Shadow Removal, State Space Models, Dual-Scale Fusion, Dynamic Scanning.
\end{IEEEkeywords}

\section{Introduction}

Shadows are common in natural images and often lead to visually distracting artifacts that adversely affect both human perception \cite{human1,human2} and the performance of high-level vision tasks \cite{girshick2014rich,SGG,VQA}. 
The goal of shadow removal is to restore clean, shadow-free images from inputs that are only partially degraded. Unlike other low-level restoration tasks, such as deblurring \cite{image_deblurring,image_deblu2} or denoising \cite{image_denoising,zhu2025mambaDenoising}, which typically assume globally distributed degradation, shadow degradation is inherently spatially localized and non-uniform. This unique nature presents both opportunities and challenges: 
% On the one hand, non-shadow regions provide valuable references for recovering shadowed areas; On the other hand, the transformations required to correct shadow regions often differ significantly from those applied to well-lit regions, making uniform processing strategies ineffective.
while non-shadow regions offer useful guidance for recovery, the corrections needed for shadow regions often differ greatly from those for well-lit regions, making uniform processing strategies ineffective.

% \begin{figure}[!t]
% \centering
% \includegraphics[width=2.5in]{fig1}
% \caption{Simulation results for the network.}
% \label{fig_1}
% \end{figure}

\begin{figure}[t]
  \centering
  \includegraphics[width=\linewidth]{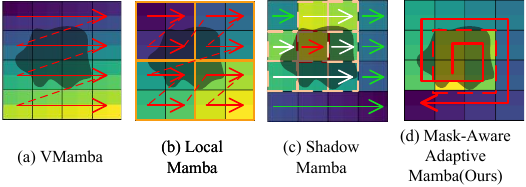}
  \caption{Illustrates four different Mamba scanning strategies on the SRD dataset: (a) VMamba \cite{liu2024vmamba}, sequential scanning with PSNR = 34.70; (b) LocalMamba \cite{huang2024localmamba}, global scanning across windows with PSNR = 29.25; (c) ShadowMamba \cite{zhu2024shadowmamba}, a Mamba mechanism designed for shadow removal with PSNR = 33.63; (d) Ours, dynamically determines the scanning starting point based on the shadow mask and uses a breadth-first search algorithm to select the next patch to scan with PSNR = \textbf{35.13}.}

  \label{fig_1}
\end{figure}

Existing deep learning-based methods have made considerable progress by leveraging convolutional neural networks (CNNs) \cite{cnn} and, more recently, transformer architectures \cite{vaswani2017attention}. CNN-based approaches \cite{chen2021DHAN,li2023cnnshadow,qu2017deshadownet} often exploit local contextual cues but lack the capacity to capture long-range dependencies, which are crucial for understanding global scene structure and effectively transferring illumination and texture information across regions.
Transformer-based models \cite{Transformer_based,Transformer_b2}, in contrast, excel at capturing global interactions but often suffer from high computational cost, and their attention mechanisms may lack adaptability to the spatial heterogeneity introduced by shadows.
% excel at modeling global interactions but often do so at the cost of computational efficiency, and their attention mechanisms may not be sufficiently adaptive to the spatial heterogeneity introduced by shadows.
More critically, both paradigms often apply a shared processing strategy across the entire image, overlooking the unique transformation requirements of shadowed versus non-shadowed regions.
Recent attempts to address this limitation have explored multi-scale processing~\cite{multi-scale,lu2023multiscale} or guided attention mechanisms~\cite{fukui2019attention}. However, these typically suffer from suboptimal fusion schemes or fail to propagate context based on transformation similarity.

Recently, the State Space Model (SSM) architecture Mamba has attracted widespread attention for simultaneously offering long-range modeling capability and linear computational complexity. As shown in Figure~\ref{fig_1}, for vision applications, researchers have explored various Mamba scanning strategies, including rigid row/column scans\cite{liu2024vmamba}, fixed local-window scans\cite{huang2024localmamba}, and region-grouped scans\cite{zhu2024shadowmamba}. 
However, these methods struggle with issues like disrupting spatial adjacency or failing to capture global dependencies effectively. 
As a result, a more tailored framework that integrates transformation-aware adaptation and efficient context propagation remains highly desirable.
% Yet these designs still limit cross-region information exchange: strict row–column serialization disrupts spatially adjacent relationships, fixed windows hamper the capture of global dependencies, and region grouping can weaken smooth transitions across boundaries. When directly applied to shadow removal, such limitations often lead to boundary artifacts or color inconsistencies.

% To this end, we leverage the Mamba architecture, a recently proposed state space model \cite{gu2021_S4} that offers efficient and scalable long-range sequence modeling, as the backbone of our design. Unlike transformers, Mamba enables selective and lightweight context propagation, which is particularly well-suited for the spatially localized yet globally informed nature of shadow removal.
Based on this insight, we propose D2-Mamba, a novel Mamba-based network architecture specifically designed for shadow removal. Our design centers around two key innovations: (1) a Dual-Scale Fusion Mamba Block (DFMB) that fuses original and low-resolution features to enhance multi-scale representation and reduce boundary artifacts; and (2) a Dual-Path Mamba Group (DPMG) module that captures global structure using horizontal scanning while incorporating a mask-aware adaptive scanning strategy to preserve fine details and structural continuity across shadow and non-shadow regions. This strategy ensures that patches from the same region (shadow or non-shadow) are more closely grouped in the Mamba sequence, enhancing detail restoration. These two scanning mechanisms complement each other, improving the robustness of the entire module. As a result, the model can effectively capture shadow boundary information, reduce artifacts, and significantly enhance shadow removal quality.
Extensive experiments on multiple benchmarks show that our method outperforms state-of-the-art models, while also delivering notable gains in efficiency and visual fidelity, validating its effectiveness for real-world shadow removal.
% Extensive experiments on several benchmarks demonstrate that our method not only outperforms existing state-of-the-art models in terms of PSNR, SSIM, and RMSE, but also achieves notable improvements in efficiency and visual fidelity, validating its effectiveness for real-world shadow removal tasks.

Our main contributions are summarized as follows:
\begin{itemize} 
\item We propose a Dual-Path Mamba Group (DPMG), which integrates horizontal scanning and a mask-aware adaptive scanning strategy to enhance both global consistency and fine-grained detail recovery in shadow regions. 
\item We introduce a Dual-Scale Fusion Mamba Block (DFMB) that facilitates cross-layer information flow and feature fusion by leveraging scale-dependent representational differences in shadow regions, improving restoration quality in shadow areas.
% interaction and feature fusion by exploiting the representational differences of shadow regions across feature scales, effectively improving restoration quality in shadow areas.
\item Extensive experiments on several datasets demonstrate that our method achieves state-of-the-art performance, while offering improved inference speed and computational efficiency.
% while offering advantages in inference speed and computational efficiency.
\end{itemize}

\section{Related Work}

% \textbf{Shadow Removal.} 
\subsection{Shadow Removal}
Early deep learning-based shadow removal methods mainly relied on CNNs to map shadowed images to shadow-free counterparts. For example, DeshadowNet \cite{qu2017deshadownet} uses three jointly trained subnetworks to integrate semantic information for improved restoration. T-CGAN \cite{wang2018T-CGAN} leverages stacked CGANs \cite{CGAN} to generate both a mask and a shadow-free image, enhancing performance through mutual task improvement. Mask-ShadowGAN \cite{hu2019mask} incorporates masks into the GAN \cite{GAN} framework, refining shadow region localization. Despite progress, CNN-based methods struggle with limited receptive fields, failing to capture long-range dependencies, particularly in complex illumination conditions.
Transformer-based architectures \cite{vaswani2017attention} alleviates these issues by capturing long-range dependencies. HomoFormer \cite{xiao2024homoformer} improves local self-attention adaptability by homogenizing shadow distributions, while ShadowFormer \cite{guo2023shadowformer} uses global attention to enhance boundary restoration. However, the high computational complexity of Transformers hinders their practical use in high-resolution tasks.
\subsection{State Space Models in Computer Vision} 
SSMs have gained attention due to low computational complexity and strong ability to model long-range dependencies.
The Structured State Space Sequence Model (S4) \cite{gu2021_S4} introduced a linear-complexity hidden state update, boosting adoption in language and vision tasks \cite{sarem2024improving, zhang2025rethinking, nasiri2024vim4path}. 
Mamba further improved this with parallel computation and a refined update scheme.
SSMs have recently been applied to low-level vision tasks such as image super-resolution \cite{lei2024dvmsrSuper-Resolution,qiao2024hiMAMBA,xiao2024frequencySuper-Resolution} and denoising \cite{lu2024lfmamba,zhu2025mambaDenoising}, showing significant improvements in computational efficiency and reconstruction quality. MambaIR \cite{vasluianu2024mambair} introduced a selective SSM approach for image restoration, enhancing reconstruction quality. LocalMamba \cite{huang2024localmamba} utilized a local window scanning strategy to improve local interactions while retaining global context. Hi-mamba \cite{qiao2024hiMAMBA} proposed a hierarchical Mamba structure for better multi-scale information fusion and feature representation.

For shadow removal, general Mamba models often fail to account for the unique characteristics of shadow images, disrupting semantic relationships at boundaries. ShadowMamba \cite{zhu2024shadowmamba} introduced a selective scanning strategy for region-based partitioning, but its rigid partitioning limits adaptability. To address these issues, we enhance cross-layer information fusion and introduce a mask-aware scanning mechanism for better structural continuity and region modeling.

\begin{figure*}[t]
\centering
\includegraphics[width=0.9\textwidth]{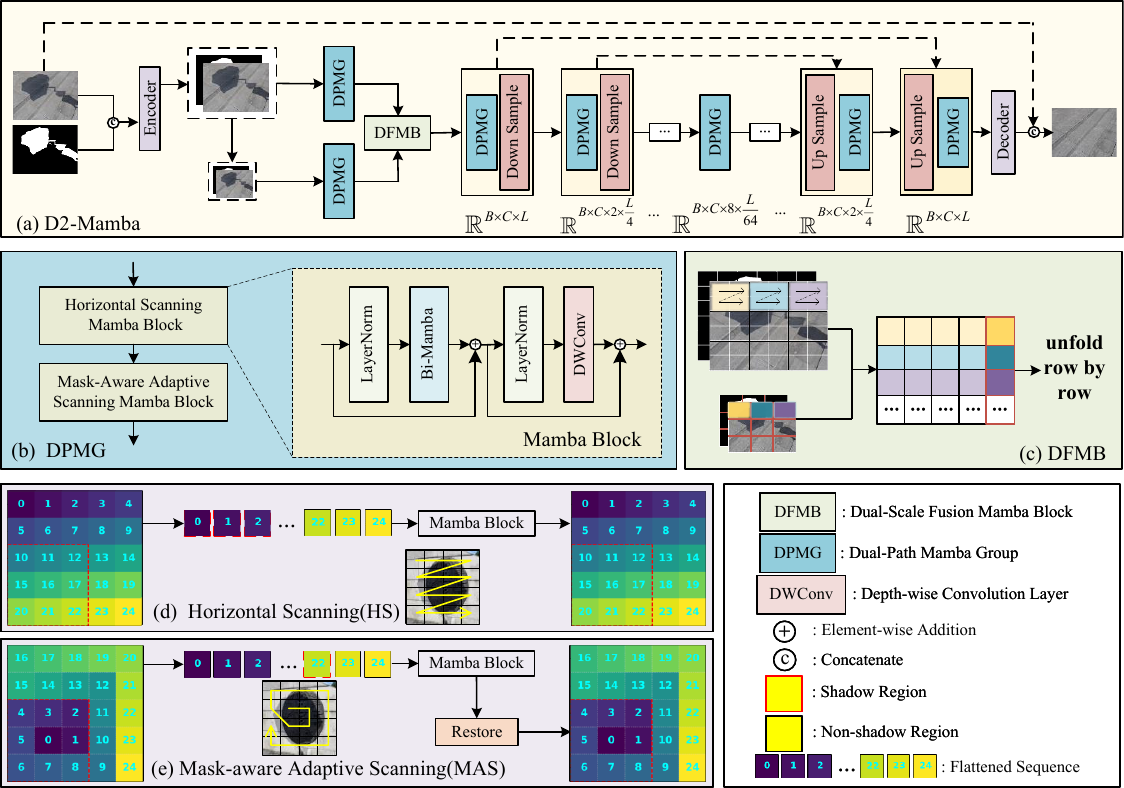} % Reduce the figure size so that it is slightly narrower than the column.
\caption{(a) An overview of our proposed D2-Mamba framework consisting of two core modules: (b) Dual-Path Mamba Group (DPMG) extracts global context information through one path using the (d) Horizontal Scanning strategy and (e) Mask-aware Adaptive scanning strategy; (c) Dual-Scale Fusion Mamba Block (DFMB) enhances the model's ability to capture multi-scale contextual information by fusing original resolution and downsampled features.}
\label{fig_2}
\end{figure*}

\section{Method}
\subsection{Preliminaries}

State Space Model (SSM) is a mathematical framework used to describe and analyze dynamic systems. 
It characterizes the internal state of a system through state variables and establishes the relationships between the state, input, and output. 
Specifically, an SSM maps a one-dimensional input sequence $x(t) \in \mathbb{R}$ to an output sequence $y(t) \in \mathbb{R}$ while propagating information through a hidden state $h(t) \in \mathbb{R}^N$. 
Its continuous-time formulation can be expressed by the following differential equations.
\begin{equation}
\begin{cases}
    h'(t) = A h(t) + B x(t) \\
    y(t) = C h(t) + D x(t)
\end{cases}
\end{equation}
where $A\!\in\!\mathbb{R}^{N \times N}$ governs the state transition, while $B\!\in\!\mathbb{R}^{N \times 1}$ and $C\!\in\!\mathbb{R}^{1 \times N}$ establish the relationships between the input and state, as well as the state and output, respectively.
The term $D$ represents the direct transmission component.
% The term $D  \in \mathbb{R}$ represents the direct transmission component.

To enable discrete sequence processing in deep learning, the continuous $A$ and $B$ are discretized into $\overline{A}$ and $\overline{B}$:
\begin{equation}
\begin{cases}
    \overline{A} = \exp(\Delta A) \\
    \overline{B} = (\Delta A)^{-1} (\exp(\Delta A) - I) \cdot \Delta B
\end{cases}
\end{equation}
Furthermore, SSM can be efficiently computed in a convolutional form. By expanding the discretized parameters into a structured convolutional kernel $\overline{K} \in \mathbb{R}^{L}$ (where $L$ represents the sequence length), the output can be expressed as the convolution of the input sequence $x$ with $\overline{K}$, where ``$\ast$'' denotes the convolution operation:
\begin{equation}
y = x * \overline{K}, \quad \overline{K} = \left( C\overline{B}, C\overline{A}\overline{B}, \dots, C\overline{A}^{L-1}\overline{B} \right).
\end{equation}
In traditional SSM, the parameters $A$, $B$, $C$, and $\Delta $ are fixed values, while Mamba builds on S4 by making them input-dependent, achieving both linear complexity and greater modeling flexibility.
% Mamba extends this by building on the S4 model, allowing these parameters to dynamically depend on the input $x$. This enhancement endows SSMs with both linear computational complexity and greater flexibility in modeling long sequences.
% However, the Mamba model further introduces the S4 Model, which allows these parameters to dynamically depend on the input $x$, this improvement provides SSM with both linear complexity and flexibility in modeling long sequences.

\subsection{Overall Architecture}

Figure~\ref{fig_2} presents an overview of the proposed D2-Mamba framework, which includes a feature encoder, a DFMB module, a UNet built with DPMG blocks, and a feature decoder.

The encoder takes the concatenated shadow image and mask as input, producing compact features $F_t \in \mathbb{R}^{L \times C}$, where $L$ denotes the spatial dimension (\textit{i.e.}, the product of image height and width) and $C$ the number of channels. These features are processed at two different scales by DFMB, which fuses them to enhance global semantics and capture shadow distributions effectively.
The fused features are subsequently processed by a UNet-like architecture composed of DPMG modules.
DPMG integrates two global scanning strategies: a horizontal Z-shaped scan and a mask-aware adaptive scan that prioritizes shadow regions. 
These strategies help the model better capture shadow structures and boundaries. The UNet-like architecture performs four downsampling and four upsampling operations with a central bottleneck, gradually compressing and restoring the spatial resolution.
Finally, the decoder generates the shadow-free output. By combining dual-scale feature fusion with complementary scanning paths, D2-Mamba enables efficient and robust removal of complex shadows.

\subsection{Dual-Path Mamba Group}

% To handle different transformations for shadow and non-shadow regions, we propose a dual-path Mamba group module with a dual scanning strategy (Figure~\ref{fig_2}(b)). A mask-aware adaptive scanning mechanism groups shadow regions, improving consistency in shadow area transformations. Simultaneously, a horizontal scanning pathway captures local continuity and context across both regions, enhancing information flow.
To effectively handle the distinct transformation requirements of shadow and non-shadow regions, we propose a dual-path Mamba group module incorporating a dual scanning strategy (Figure~\ref{fig_2}(b)). Specifically, a mask-aware adaptive scanning mechanism is employed to group and process shadow regions, ensuring consistent transformations within shadow areas. In parallel, a horizontal scanning pathway captures local continuity and contextual information across both shadow and non-shadow regions, thereby enhancing the overall information flow and integration.

As shown in Figure~\ref{fig_2}(d), the HSMB employs a row-wise scanning strategy. Given an input feature map $X \in \mathbb{R}^{C \times H \times W}$, we flatten it into a sequence $V \in \mathbb{R}^{C \times (H \times W)}$ and apply bidirectional scanning (left-to-right and right-to-left), enabling feature interactions in a single step.
Each Mamba block uses ConvMLP with two fully connected layers for feature transformation. The first layer projects features into a high-dimensional space, followed by depthwise separable convolution (DWConv) for local feature enhancement. After passing through GELU and Dropout, the features are projected back to the original dimension:
% Shadow and non-shadow regions often require different transformations. To address this, we propose a novel dual-path mamba group module that adopts a dual scanning strategy illustrated in Figure~\ref{fig_2}(b). A mask-aware adaptive scanning mechanism is introduced to group shadow regions, enabling the model to learn more consistent transformations within shadow areas. In parallel, a horizontal scanning pathway is applied to capture local continuity and neighboring context across shadow and non-shadow areas, facilitating better information flow between regions.
% Specifically, as illustrated in Figure~\ref{fig_2}(d), the Horizontal Scanning Mamba Block (HSMB) processes input features using a row-wise scanning strategy. 
% Specifically, given an input feature map $X \in \mathbb{R}^{C \times H \times W}$, we first flatten it row by row into a one-dimensional sequence $V \in \mathbb{R}^{C \times (H \times W)}$. 
% Then we adopt a bidirectional scanning mechanism to simultaneously perform forward (left-to-right) and backward (right-to-left) state updates for $V$, enabling bidirectional feature interactions within a single time step. 
% In each Mamba block, we employ ConvMLP with two fully connected layers for feature transformation.
% The first layer projects the features into a high-dimensional space, followed by a depthwise separable convolution (DWConv) to enhance local feature representation. 
% The features then pass through a GELU activation and a Dropout layer before being projected back to the original feature dimension. This process is formulated as follows:
\begin{equation}
z = \text{Dropout}(W_2(\text{GELU}(\text{DWConv}(W_1 x)))).
\end{equation}
% The Horizontal Scanning Mamba block 

% HSMB establishes long-range dependencies through its bidirectional state update mechanism, effectively capturing correlations across regions and enhancing feature representation capability.
HSMB captures long-range dependencies via a bidirectional state update, effectively modeling regional correlations and enhancing feature representation.

\begin{algorithm}[t]
\caption{Mask-Aware Adaptive Scanning Strategy}
\label{alg1}
\textbf{Input:} \( r; c \); \( sub\_r; sub\_c \); \( cx_{sub}; cy_{sub} \); \( directions\) = \{(-1,0), (1,0), (0,-1), (0, 1)\}.\\
\textbf{Output:} $path$
\begin{algorithmic}[1]

\STATE \textbf{Select the Starting Point of Non-Shadow Region A:}
% \STATE \( Visited \gets \emptyset\)
\STATE \(t,b,l,r \gets rect\_ bounds(r,c,sub\_r,sub\_c,cx_{sub},cy_{sub})\)
\STATE \( dist \gets \{ t, b, l, r \} \)
% \noindent \textbf{\textit{\textcolor{gray}{\# Calculate the distance to each edge of shadow region}}}
\noindent \textbf{\textit{\#Distance to each edge of full image}}
\STATE \( edge1 \gets choose\_closest\_edge(dist) \)
\STATE \( edge2 \gets choose\_adjacent\_edge(dist, edge1) \)
% \noindent \textbf{\textit{\textcolor{gray}{\# Determine the start corner coordinates}}}
\STATE \(start \_ A \gets (edge1,edge2)\) 
\STATE \textbf{Traverse Non\_Shadow Region A:}
\STATE \(visited \gets \{ (x, y) \mid t \leq x \leq b, l \leq y \leq r \} \) 
% \noindent \textbf{\textit{\textcolor{gray}{\# Mark the shadow region \( \mathbf{B} \) as visited}}}
\STATE \( path\_A \gets [] \) 
% \STATE \( cur \gets start\_A \) \textbf{\textit{\textcolor{gray}{\# Set the current position as the start corner}}}
\WHILE {\( \lvert visited \rvert < r \times c \)}
    \STATE \( best \gets \text{None}, best\_touch \gets -1 ,eq = [] \)
    \FOR{each \( (dr, dc) \in \text{directions} \)}
        \STATE \( nr, nc \gets cur[0] + dr, cur[1] + dc \)
        \STATE \(touch = calculate\_touch (nr, nc)\)
        \IF{\( touch > best\_touch \)}
            \STATE \( candList \gets (nr, nc), best\_touch \gets touch\)
        \ELSIF{ \(touch == best\_touch \)}
            \STATE \( candList \gets candList \cup \{(nr,nc)\} \)
        \ENDIF
    \ENDFOR
    \IF{ \( candList \text{ is None}\)}
        \STATE \(best \gets  min\_manhattan(unvisited) \)
    \ELSE
        \STATE \(best \gets select\_candlist(candList) \)
        % \FOR{\( \text{each} (dr,dc) \in directions\)}
        %     \STATE \(next \gets (cur[0]+dr,cur[1]+dc)\)
        %     \IF{\(next \in candList\)}
        %         \STATE\( best \gets next\)
        %         \STATE break
        %     \ENDIF
        % \ENDFOR
    \ENDIF
    \STATE \(cur \gets best\)
    \STATE \(visited \gets visited \cup \{cur\}\)
    \IF{\( (cur[0],cur[1]) \notin B\)}
        \STATE \(path\_A \gets path\_A \cup \{cur\}\)
    \ENDIF
\ENDWHILE
% \STATE \textbf{return} $path_A$
\STATE \textbf{Traverse Shadow Region B:}
\STATE \( start\_B \gets choose\_closest\_point\_in\_B(start\_A)\)
\STATE \( path\_ B \gets reversr(spiral\_in(start\_B))\)
\STATE \textbf{return} $path\_B \cup path\_A $

\end{algorithmic}

\end{algorithm}

\begin{table*}[t]
\begin{center}
\caption{The quantitative results of shadow removal on SRD dataset. The best and the second results are in bold and underlined.}
\label{tab_1}
\setlength{\tabcolsep}{2mm}  
\fontsize{9}{\baselineskip}\selectfont 
\begin{tabular}{l|ccc|ccc|ccc}
\toprule
\multirow{2}{*}{Method} & \multicolumn{3}{c|}{Shadow Region (S)} & \multicolumn{3}{c|}{Non-Shadow Region (NS)} & \multicolumn{3}{c}{All Image (ALL)} \\
                        & PSNR $\uparrow$ & SSIM $\uparrow$ & RMSE $\downarrow$ & PSNR $\uparrow$ & SSIM $\uparrow$ & RMSE $\downarrow$ & PSNR $\uparrow$ & SSIM $\uparrow$ & RMSE $\downarrow$ \\
\midrule
% Guo et al. \cite{guo2012paired} & - & - & 29.89 & - & - & 6.47 & - & - & 12.6 \\
DeshadowNet \cite{qu2017deshadownet} & - & - & 11.78 & - & - & 4.84 & - & - & 6.64 \\
% \hline
% DSC \cite{hu2018dsc} & 30.65 & 0.960 & 8.62 & 31.94 & 0.965 & 4.41 & 27.76 & 0.903 & 5.71 \\
DHAN \cite{chen2021DHAN} & 33.67 & 0.978 & 8.94 & 34.79 & 0.979 & 4.80 & 30.51 & 0.949 & 5.67 \\
% \hline
% Fu et al. \cite{fu2021autoFuetal} & 32.26 & 0.966 & 8.55 & 31.87 & 0.945 & 5.74 & 28.40 & 0.893 & 6.50 \\
BMNet \cite{zhu2022BMNet} & 35.05 & 0.981 & 6.61 & 36.02 & 0.982 & 3.61 & 31.69 & 0.956 & 4.46 \\
% \hline
SGNet \cite{wan2022SGNet} & 33.76 & 0.979 & 7.45 & 36.48 & 0.984 & 3.05 & 31.39 & 0.960 & 4.23 \\
% \hline
ShadowFormer-L \cite{guo2023shadowformer} & 36.91 & \textbf{0.989} & 5.90 & 36.22 & \textbf{0.989} & 3.44 & 32.90 & 0.958 & 4.04 \\
% \hline
ShadowDiffusion \cite{guo2023shadowdiffusion} & \underline{38.72} & \underline{0.987} & 4.98 & 37.78 & 0.985 & 3.44 & 34.73 & \underline{0.970} & 3.63 \\
% \hline
HomoFormer \cite{xiao2024homoformer} & \textbf{38.81} & \underline{0.987} & \textbf{4.25} & \underline{39.45} & \underline{0.988} & \underline{2.85} & \underline{35.37} & \textbf{0.972} & \underline{3.33} \\
% \hline
Liu et al. \cite{liu2024RR} & 36.51 & 0.983 & 5.49 & 37.71 & 0.986 & 3.00 & 33.48 & 0.967 & 3.66 \\
% \hline
ShadowMamba \cite{zhu2024shadowmamba} & 37.29 & 0.986 & 5.81 & 37.52 & 0.985 & 3.13 & 33.63 & 0.965 & 3.87 \\
% \hline
% Diff-shadow  \cite{luo2025diff_shadow} & 37.91 & 0.988 & 1.81 & 39.49 & 0.990 & 1.65 & 34.93 & 0.977 & 2.52 \\
ShadowGAN-Former~\cite{hu2025shadowgan} & 35.24 & 0.981 & 6.578 & 36.12 & 0.981 & 3.54 & 31.73 & 0.956 & 4.39 \\
% \hline
ShadowMaskFormer \cite{li2025shadowmaskformer} & 37.71 & 0.988 & 5.55 & 38.23 & 0.984 & 2.98 & 34.43 & 0.968 & 3.64 \\
% \hline
MSRDNet \cite{huang2025MSRDNet} & 35.43 & 0.984 & 5.98 & 36.23 & 0.989 & 3.38 & 32.17 & 0.965 & 4.09 \\
% \hline
OmniSR \cite{omnisr} & - & - & - & - & - & - & 34.56  & 0.977 & - \\
\midrule
D2-Mamba (Ours) & 38.46 & 0.981 & \underline{4.38} & \textbf{40.02} & 0.983 & \textbf{2.77} & \textbf{35.41} & 0.958 & \textbf{3.31} \\
\bottomrule
\end{tabular}
\end{center}
\end{table*}

\begin{table*}[t]
\begin{center}
\caption{The quantitative results of shadow removal on the ISTD+ dataset.}
\label{tab_3}
\setlength{\tabcolsep}{2mm}
\fontsize{9}{\baselineskip}\selectfont  

\begin{tabular}{l|ccc|ccc|ccc}
\toprule
\multirow{2}{*}{Method} & \multicolumn{3}{c|}{Shadow Region (S)} & \multicolumn{3}{c|}{Non-Shadow Region (NS)} & \multicolumn{3}{c}{All Image (ALL)} \\
                        & PSNR $\uparrow$ & SSIM $\uparrow$ & RMSE $\downarrow$ & PSNR $\uparrow$ & SSIM $\uparrow$ & RMSE $\downarrow$ & PSNR $\uparrow$ & SSIM $\uparrow$ & RMSE $\downarrow$ \\
\midrule
DHAN \cite{chen2021DHAN} & 32.92 & 0.988 & 9.6 & 27.15 & 0.971 & 7.4 & 25.66 & 0.956 & 7.8 \\
% \hline
SGNet \cite{wan2022SGNet} & 36.79 & 0.990 & 5.9 & 35.57 & 0.977 & 2.9 & 32.45 & 0.962 & 3.4 \\
% \hline
BMNet \cite{zhu2022BMNet} & 37.87 & \underline{0.991} & 5.8 & 37.51 & \textbf{0.985} & 2.4 & 33.98 & 0.972 & 3.0 \\
% \hline
ShadowFormer-L \cite{guo2023shadowformer} & 39.67 & - & 5.2 & 38.82 & - & 2.3 & 35.46 & - & 2.8 \\
% \hline
ShadowDiffusion \cite{guo2023shadowdiffusion} & \underline{39.8} & - & \underline{4.9} & \underline{38.90} & - & \underline{2.3}  & \underline{35.72} & -& \underline{2.7} \\
% \hline
HomoFormer \cite{xiao2024homoformer} & 39.49 & \underline{0.991} & \underline{4.9} & 38.64 & 0.982 & 2.3 & 35.31 & 0.969 & \textbf{2.6} \\
Liu et al. \cite{liu2024RR} & 38.04 & 0.990 & 5.7 & 39.15 & \underline{0.984} & 2.3 & 34.96 & 0.968 & 2.9 \\
% \hline
ShadowMamba \cite{zhu2024shadowmamba} & - & - & 5.8 & - & - & \underline{2.3} & - & - & 2.8 \\
% \hline
% Diff-shadow  \cite{luo2025diff_shadow} & 40.63 & 0.993 & 1.3 & 39.69 & 0.989 & 1.6 & 36.47 & 0.979 & 2.1 \\
MSRDNet \cite{huang2025MSRDNet} & 38.93 & \underline{0.991} & 5.5 & 38.49 & \textbf{0.985} & 2.4 & 34.94 & \underline{0.972} & 2.9 \\
% \hline
OmniSR \cite{omnisr} & - & - & - & - & - & - &  34.20  & \textbf{0.973} & - \\
% \hline
\midrule
D2-Mamba (Ours) & \textbf{40.22} & \textbf{0.992} & \textbf{4.6} & \textbf{38.95} & \textbf{0.985} & \textbf{2.2} & \textbf{35.84} & \textbf{0.973} & \textbf{2.6} \\
\bottomrule
\end{tabular}

\end{center}
\end{table*}

\noindent\textbf{Mask-aware Adaptive Scanning.}
After establishing long-range dependencies with HSMB, we propose a novel Mask-aware Adaptive Scanning (MAS) to improve structural continuity and fine-grained region modeling.
In the shadow regions, we employ a spiral scan, while for the non-shadow areas, we adopt the Greedy Boundary-contact Search (GBS) strategy to ensure scan continuity. The detailed search process of the proposed mask-aware adaptive scanning strategy is outlined in Algorithm~\ref{alg1}.

For clarity, we use the example shown in Figure \ref{fig_2}(e). In this case, we first perform spatial partitioning based on the input feature map $x \in \mathbb{R}^{C \times H \times W}$ and the shadow mask $m \in \mathbb{R}^{1 \times H \times W}$, dividing the image into $\frac{H}{8} \times \frac{W}{8}$ sub-blocks of size $s \times s$. 
The average mask value ${\overline m_{i,j}}$ or each sub-block is then computed to determine whether it belongs to the shadow region. 
At this point, the image has $r$ rows and $c$ columns, and the shadow region has $\text{sub\_r}$ rows and $\text{sub\_c}$ columns. We then compute the center position of the shadow block as $(cx_{\text{sub}}, cy_{\text{sub}})$ and calculate the distance to the image boundaries as follows:
\begin{equation}
\begin{cases}
\textit{t}&= \max\left(0,\, cx_{\text{sub}} - \frac{{\text{sub\_r} - 1}}{2}\right) \\
\textit{b}&= \min\left(\text{r} - 1, cx_{\text{sub}} + \left\lfloor \frac{\text{sub\_r}}{2} \right\rfloor\right) \\
\textit{l}&=  \max\left(0,\, cy_{\text{sub}} - \frac{{\text{sub\_c} - 1}}{2}\right) \\
\textit{r}&= \min\left(\text{c} - 1, cy_{\text{sub}} + \left\lfloor \frac{\text{sub\_c} - 1}{2} \right\rfloor\right) \\
\end{cases}
\end{equation}

Using these distances, we set the starting point for the non-shadow region as $start\_A$. Next, the GBS strategy is employed to find the next grid with the most boundary contacts to the previously visited grid, which is then added to the path $path\_A$. 
For the shadow region, we select the closest point to the starting point of the non-shadow area as the initial point $start\_B$ and apply a spiral path inward, reversing it to obtain $path\_B$. The final sequence is 
$path=[path\_B, path\_A]$, and we apply the Mamba and ConMLP models to perform feature transformation and modeling.

\subsection{Dual-Scale Fusion Mamba Block}

\begin{table*}[t]
\begin{center}
\caption{The quantitative results of shadow removal using our models and recent methods on the ISTD dataset.}
\label{tab_2}
\setlength{\tabcolsep}{2mm}
\fontsize{9}{\baselineskip}\selectfont 
\begin{tabular}{l|ccc|ccc|ccc}
\toprule
\multirow{2}{*}{Method} & \multicolumn{3}{c|}{Shadow Region (S)} & \multicolumn{3}{c|}{Non-Shadow Region (NS)} & \multicolumn{3}{c}{All Image (ALL)} \\
                        & PSNR $\uparrow$ & SSIM $\uparrow$ & RMSE $\downarrow$ & PSNR $\uparrow$ & SSIM $\uparrow$ & RMSE $\downarrow$ & PSNR $\uparrow$ & SSIM $\uparrow$ & RMSE $\downarrow$ \\
\midrule
% DSC \cite{hu2018dsc} & 34.64 & 0.984 & 8.72 & 31.26 & 0.969 & 5.04 & 29.00 & 0.944 & 5.59 \\
DHAN \cite{chen2021DHAN} & 34.65 & 0.983 & 8.26 & 29.81 & 0.937 & 5.56 & 28.15 & 0.913 & 6.37 \\
% \hline
% SP+M+I-Net \cite{le2021physicsSPMI} & 32.89 & 0.986 & 10.84 & 26.11 & 0.965 & 7.40 & 25.01 & 0.948 & 7.96 \\
BMNet \cite{zhu2022BMNet} & 35.61 & 0.988 & 7.60 & 32.80 & 0.976 & 4.59 & 30.28	& 0.959	& 5.02 \\
% \hline
ShadowFormer-S \cite{guo2023shadowformer} & 37.99 & 0.990 & 6.16 & \underline{33.89} & 0.980 & \underline{3.90} & 31.81 & 0.967 & 4.27 \\
% \hline
ShadowDiffusion \cite{guo2023shadowdiffusion} & \textbf{40.15} & \textbf{0.994} & \textbf{4.13} & 33.70 & 0.977 & 4.14 & \textbf{32.33} & \textbf{0.969}  & \underline{4.12} \\
% \hline
HomoFormer \cite{xiao2024homoformer} & 39.07 & 0.991 & 5.64 & 33.45 & \underline{0.979} & 4.10 & 31.82 & \underline{0.968} & 4.33 \\
% \hline
ShadowMamba \cite{zhu2024shadowmamba} & 38.01 & 0.990 & 5.95 & 33.78 & 0.974 & 3.90 & 31.79 & 0.960 & 4.23 \\
% \hline
% Diff-shadow  \cite{luo2025diff_shadow} & 37.91 & 0.988 & 1.81 & 39.49 & 0.990 & 1.65 & 34.93 & 0.977 & 2.52 \\
ShadowMaskFormer \cite{li2025shadowmaskformer} & - & - & 6.08 & - & - & 3.86 & - & - & 4.23 \\
% \hline
ShadowGAN-Former \cite{hu2025shadowgan} & 36.64 & 0.990 & 6.82 & 33.60 & 0.980 & 4.00 & 30.97 & 0.967 & 4.46 \\
% \hline
MSRDNet \cite{huang2025MSRDNet} & 37.36 & 0.989 & 6.38 & 33.38 & 0.980 & 4.09 & 31.16 & 0.965 & 4.48 \\
% \hline
OmniSR \cite{omnisr} & - & - & - & - & - & - & 31.56 &  0.965 & - \\
\midrule
D2-Mamba (Ours) & \underline{39.14} & \underline{0.992} & \underline{5.61} & \textbf{34.10} & \textbf{0.981} & \textbf{3.81} & \underline{32.18} & \textbf{0.969} & \textbf{4.06} \\
\bottomrule
\end{tabular}
\end{center}
\end{table*}

\begin{figure*}[t]
     \centering
     \includegraphics[width=0.9\textwidth]{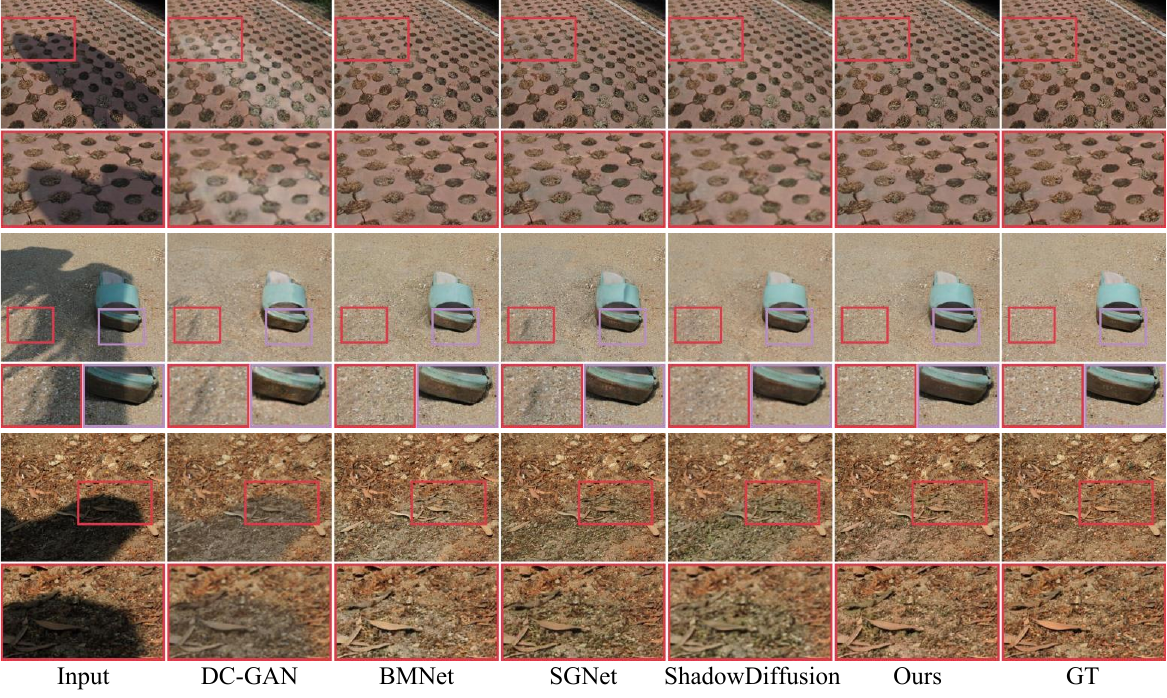}
    \caption{Visual comparisons with state-of-the-art methods on the SRD dataset.
    % According to the interaction between object pairs, we can divide them into interaction boxes and non-interaction boxes. 
    % In these two cases, according to the relative position relationship between object pairs, they can be further divided into six situations.
    % Among them, the red and blue solid boxes are object boxes, and the solid boxes corresponding to the yellow areas are our designed cross sub-boxes.
    }
    \label{fig_srd}
\end{figure*}

Previous studies~\cite{ren2024mambacross,guo2023shadowformer} demonstrate that cross-scale information is crucial for achieving consistent brightness and color in shadow removal. To further incorporate cross-scale information into the Mamba framework, we draw inspiration from the position-aligned cross-scale scanning strategy proposed by MambaCSR~\cite{ren2024mambacross}. Building on this idea, we design an improved cross-scale feature fusion strategy to leverage multi-scale features and enhance the restoration capability effectively.

% \begin{figure*}[!t]
%     \centering
%     \begin{minipage}[t]{0.48\textwidth}
%         \centering
%         \includegraphics[width=\textwidth]{SRD_crop}
%         \caption{Performance comparison of different distillation CIL methods.}
%         \label{fig:fig4}
%     \end{minipage}
%     \hfill
%     \begin{minipage}[t]{0.48\textwidth}
%         \centering
%         \includegraphics[width=\textwidth]{SRD_crop}
%         \caption{Performance comparison under different batch sizes.}
%         \label{fig:fig5}
%     \end{minipage}
% \end{figure*}

First, the input shadow image $I_s$ and its binary shadow mask $M$ are fed into the encoder to extract the initial feature:
\begin{equation}
F_{\text{input}} = \text{Flatten}\left( \text{LeakyReLU}\left( \text{Conv2d}(\text{Concat}(I_s, M)) \right) \right).
\end{equation}

To incorporate richer information, we apply bilinear interpolation to downsample the original feature map, obtaining a lower-resolution representation $F_{\text{down}}$.
Both $F_{\text{input}}$ and $F_{\text{down}}$ are subsequently processed by DPMG to model global and region-level dependencies from different shadow areas.

For feature fusion, we adopt a position-aligned unfolding strategy. 
As shown in Figure~\ref{fig_2}(c), for any pixel $x_d(i,j)$ in the downsampled feature map $F_{\text{down}}$, we extract its four spatially aligned neighbors from the original feature map $F_{\text{input}}$:$x(2i, 2j), x(2i+1, 2j), x(2i, 2j+1), x(2i+1, 2j+1)$, which are then grouped into a sequential token unit:
% \begin{equation}
% S_{i,j} = [x(2i,2j), x(2i+1,2j), x(2i,2j+1), x(2i+1,2j+1), x_d(i,j)].
% \end{equation}
\begin{equation}
\begin{aligned}
S_{i,j} &= [ x(2i,2j), x(2i+1,2j), x(2i,2j+1), \\
         &\quad x(2i+1,2j+1),x_d(i,j)], i\in [1, H/2], j \in [1, W/2].\\
\end{aligned}
\end{equation}

Next, we perform a row-wise scan of the fused feature map $S_{i,j}$, unfolding each five-element group in a sequential order to generate a 1D sequence $S$.
This sequence is fed into the fusion mamba block to obtain $F_{\text{fused}}$ for fine-grained multi-scale feature interaction.

The fused features $F_{\text{fused}}$ are then forwarded to the subsequent U-Net structure for further processing.
During the downsampling stage, the U-Net stores the intermediate features generated by the Mamba module, \textit{i.e.}, $x_\text{s} \in \mathbb{R}^{C \times H/s \times W/s}$, which are later upsampled and aligned back to the original spatial resolution $x' \in \mathbb{R}^{C \times H \times W}$.
These features are fused with corresponding encoder features in the upsampling path, allowing the network to better preserve structure and detail for more refined shadow removal.

\section{Experiments}

\subsection{Experimental Setups}

\noindent\textbf{Datasets.}
We conduct experiments on three widely used shadow removal benchmarks: SRD~\cite{qu2017deshadownet}, ISTD~\cite{wang2018T-CGAN}, and ISTD+\cite{le2019shadow}. The SRD dataset contains 3,088 shadow/shadow-free image pairs. Since it lacks ground-truth shadow masks, we use predicted masks from DHAN \cite{chen2021DHAN} as substitutes. We follow the standard split with 2,680 pairs for training and 408 for testing. The ISTD dataset provides 1,870 triplets, each consisting of a shadow image, its mask, and a corresponding shadow-free image, divided into 1,330 training and 540 testing samples. ISTD+ is an improved version of ISTD, in which illumination inconsistencies between shadow and shadow-free images have been corrected through post-processing techniques.

\noindent\textbf{Implementation Details.}
Our method is optimized using the Adam optimizer \cite{kingma2014adam} and implemented in the PyTorch framework. All experiments were conducted on a single NVIDIA RTX 4090 GPU. The initial learning rate for training is set to $2 \times 10^{-4}$, with momentum parameters $\beta_1$ and $\beta_2$ set to 0.9 and 0.999, respectively, and a batch size of 4. To optimize the training process, a cosine annealing strategy is employed to gradually reduce the learning rate, ultimately converging to $1 \times 10^{-6}$.

%%%%%%------------放到补充材料里面--------------------
\noindent\textbf{Evaluation Metrics.}
To enable quantitative comparisons with existing methods, we follow previous studies \cite{hu2018dsc,fu2021autoFuetal,le2021physicsSPMI, zhu2022BMNet,wan2022SGNet} and adopt RMSE in the LAB color space to evaluate shadow removal performance. This metric measures the difference between the predicted shadow-free images and the ground truth, where a lower RMSE indicates better restoration quality. In addition, to assess performance in the RGB color space, we also employ PSNR and SSIM as evaluation metrics.
All images used for evaluation are uniformly resized to 256×256 to ensure consistency across comparisons.

\begin{figure*}[t]
     \centering
     \includegraphics[width=0.9\textwidth]{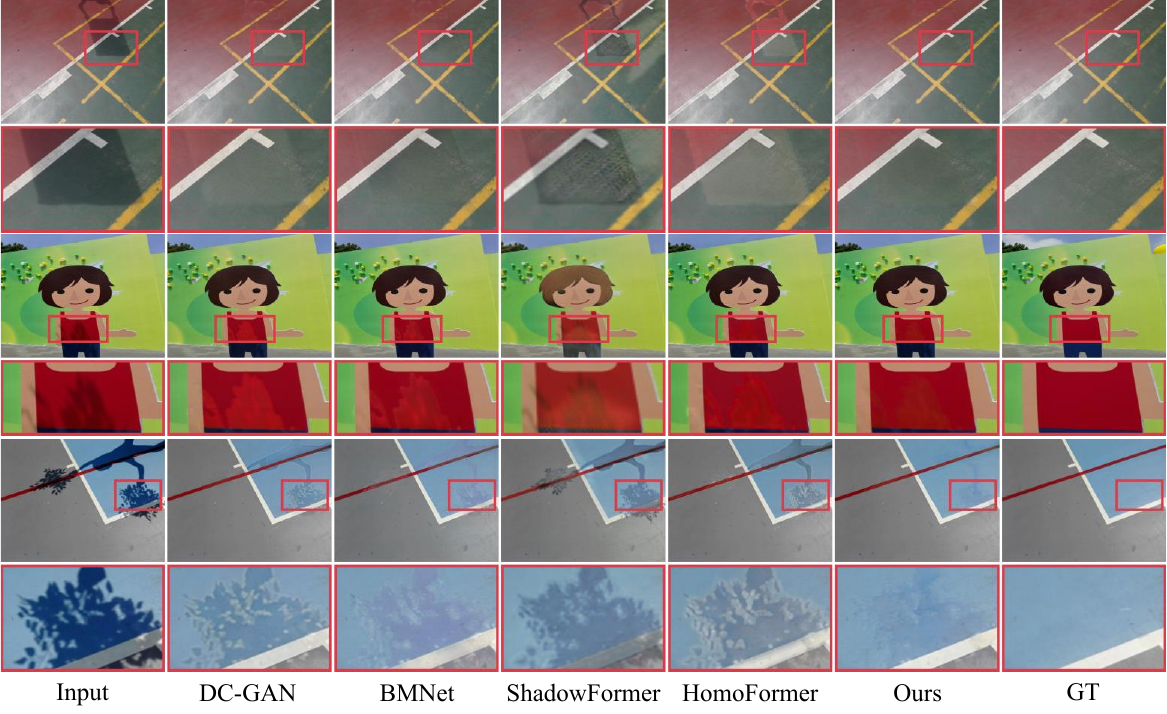}
    \caption{Visual comparisons with state-of-the-art methods on the ISTD+ dataset.
    % According to the interaction between object pairs, we can divide them into interaction boxes and non-interaction boxes. 
    % In these two cases, according to the relative position relationship between object pairs, they can be further divided into six situations.
    % Among them, the red and blue solid boxes are object boxes, and the solid boxes corresponding to the yellow areas are our designed cross sub-boxes.
    }
    \label{fig_aistd}
\end{figure*}

\begin{figure}[t]
     \centering
      \includegraphics[width=0.48\textwidth, trim=5 0 0 0, clip]{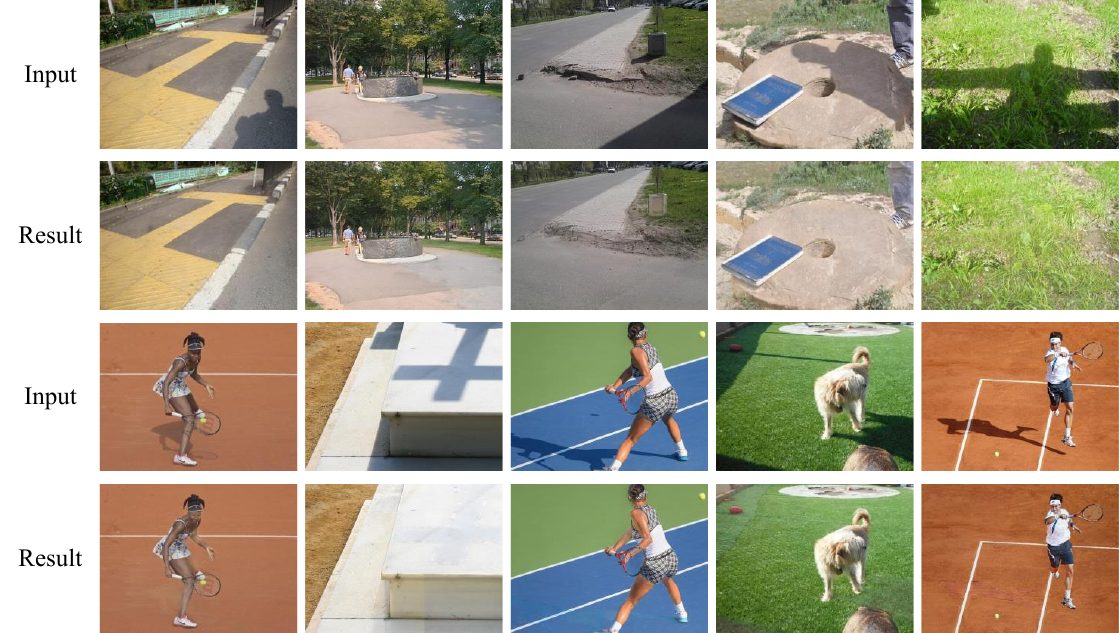}
    \caption{Shadow removal results of the proposed D2-Mamba on the SBU Shadow Dataset. Each pair shows the input shadow image and the corresponding shadow-free result produced by our model. The results demonstrate the model’s ability to handle complex shadow patterns under diverse real-world indoor lighting conditions.}
    \label{fig_6}
\end{figure}

\begin{figure*}[t]
     \centering
      \includegraphics[width=0.9\textwidth]{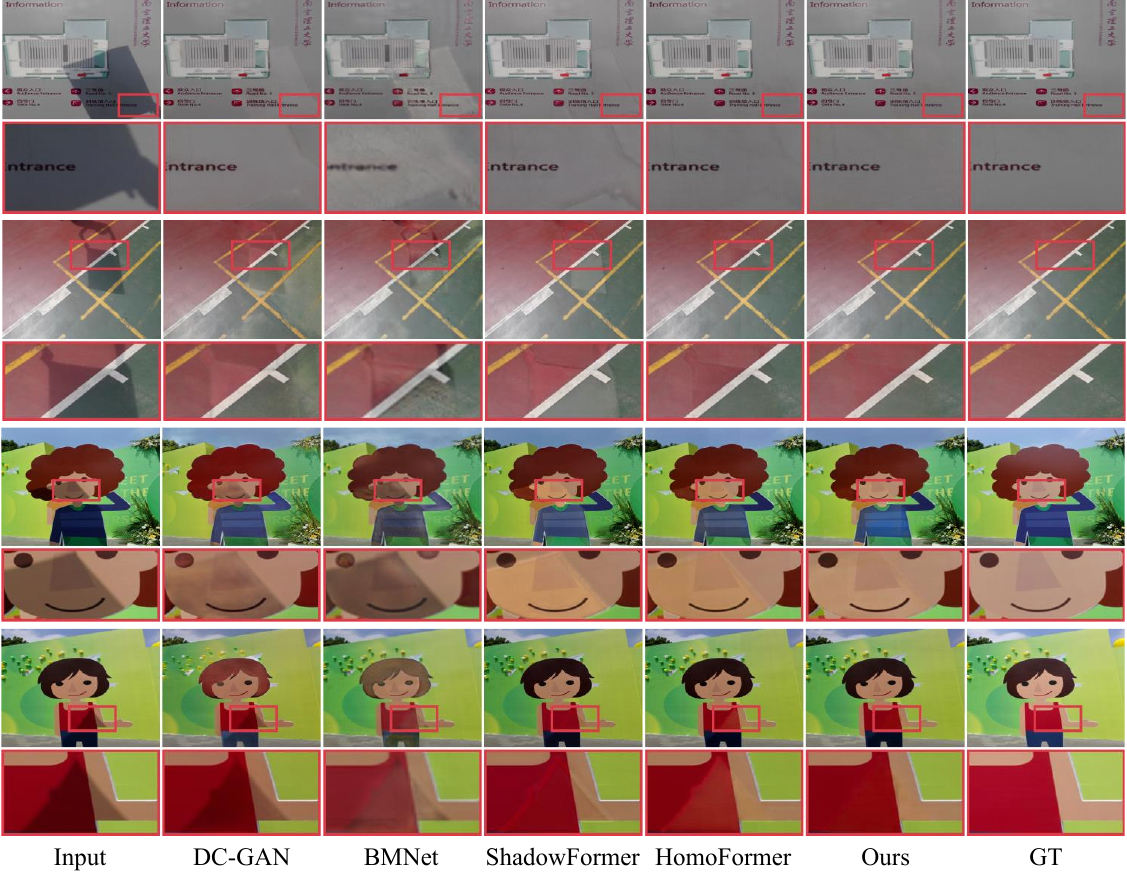}
    \caption{Visual examples of the shadow removal results on the ISTD dataset.}
    \label{fig_istd}
\end{figure*}

\subsection{Comparison with State-of-the-Art}

We compare the proposed model with popular or state-of-the-art shadow removal methods.
% including DeshadowNet \cite{qu2017deshadownet}, DSC \cite{hu2018dsc}, DHAN \cite{chen2021DHAN}, Fu et al. \cite{fu2021autoFuetal}, SP+M+I-Net \cite{le2021physicsSPMI}, BMNet \cite{zhu2022BMNet}, SGNet \cite{wan2022SGNet}, ShadowFormer \cite{guo2023shadowformer}, ShadowDiffusion \cite{guo2023shadowdiffusion}, HomoFormer \cite{xiao2024homoformer}, Liu et al. \cite{liu2024RR}, ShadowGAN-Former \cite{hu2025shadowgan}, MSRDNet \cite{huang2025MSRDNet}, and ShadowMamba \cite{zhu2024shadowmamba}.
Results of all comparison methods are either obtained from their original publications or reproduced using official implementations

\noindent\textbf{Quantitative evaluation.} Table~\ref{tab_1}, Table~\ref{tab_3} and Table~\ref{tab_2} present the quantitative evaluation results over SRD, ISTD+ and ISTD, respectively. Across all three datasets, our method consistently outperforms or matches existing approaches in multiple evaluation metrics, demonstrating its effectiveness and robustness in shadow removal.

On the SRD dataset, which includes complex scenes and textures, diffusion-based methods perform well due to their generative capacity for modeling intricate details. However, these methods suffer from long inference times and limited scalability for high-resolution inputs. In contrast, our Mamba-based regression model, equipped with accurate cross-scale fusion and similarity-aware grouping, achieves competitive or superior performance with significantly fewer inference steps, making it more efficient and practical for real-world deployment.
On the ISTD+ dataset, our method leads in all metrics, with notable improvements in both the shadow and non-shadow regions. We achieve the highest PSNR, SSIM, and the lowest RMSE, outpacing state-of-the-art methods like ShadowDiffusion and ShadowMamba. This demonstrates the robustness of our approach in handling varied shadow types and complex lighting conditions.
On the ISTD dataset, we maintain top-tier performance, outperforming or matching existing methods like ShadowFormer and ShadowDiffusion. Our approach excels in shadow regions (S) and overall image quality (ALL), with a balanced trade-off between high PSNR and SSIM scores while keeping the RMSE low.
Our D2-Mamba model offers superior shadow removal quality across three datasets, balancing efficiency with high performance.

\noindent\textbf{Qualitative evaluation.}
To further validate the effectiveness of our proposed D2-Mamba framework, we provide qualitative comparisons on the SRD, ISTD+, and ISTD datasets, which cover diverse scenes with varying shadow shapes, densities, and lighting conditions. These datasets allow for a comprehensive evaluation of the model’s robustness.

Figures~\ref{fig_srd},~\ref{fig_aistd}, and~\ref{fig_istd} clearly demonstrate that our method consistently outperforms other approaches in terms of shadow boundary smoothness and artifact removal. In the SRD and ISTD+ datasets, while some methods produce shadow-recovered regions that appear noticeably brighter or darker than their surroundings (e.g., the first and third examples in Figure~\ref{fig_srd}), or struggle with boundary artifacts (e.g., the second and fourth examples), our approach effectively removes residual shadows, restoring natural luminance and color for improved global consistency and visual realism. Additionally, challenging scenes with high-saturation areas (Figure~\ref{fig_aistd}) often lead to color inconsistencies in competing methods, whereas our method maintains color integrity across both shadow and non-shadow regions.On the ISTD dataset, D2-Mamba demonstrates strong performance in preserving texture details and structural consistency. Compared methods often produce visible residual artifacts, particularly in complex shadow regions, where shadows fail to blend seamlessly with the background or suffer from significant color distortions. In contrast, our method achieves more accurate and visually coherent shadow removal, producing natural-looking results with fewer artifacts. The shadow areas are smoothly integrated into the surrounding context, preserving the global semantic structure while accurately modeling local shadow variations.

\subsection{Real-World Shadow Removal Performance}

To further evaluate the adaptability of the proposed D2-Mamba model in complex scenarios, we conducted additional qualitative experiments on the SBU Dataset. This dataset presents more challenging conditions, containing diverse indoor lighting environments with complex shadow distributions cast by objects such as people and furniture.
Figure~\ref{fig_6} illustrates several shadow removal examples produced by our model on selected images from the SBU dataset. These results demonstrate that D2-Mamba not only performs well on standard benchmark datasets but also exhibits strong generalization ability and robustness, making it suitable for shadow removal tasks in more diverse and dynamic real-world environments.

% Modification
\begin{table*}
\begin{center}
\caption{Ablation study of the proposed D2-Mamba on the SRD dataset.}
\label{tab_4}
\setlength{\tabcolsep}{1mm}
% \fontsize{9}{\baselineskip}\selectfont  
\resizebox{0.8\textwidth}{!}{
  \begin{tabular}{c|cc|ccc|ccc|ccc}
    \toprule
    \multirow{2}{*}{DFMB} & \multicolumn{2}{c|}{DPMG} & \multicolumn{3}{c|}{Shadow Region (S)} & \multicolumn{3}{c|}{Non-Shadow Region (NS)} & \multicolumn{3}{c}{All Image (ALL)} \\
                       &HS&MAS& \text{$\text{PSNR}$} $\uparrow$ & \text{$\text{SSIM}$} $\uparrow$ & \text{$\text{RMSE}$} $\downarrow$ & \text{$\text{PSNR}$} $\uparrow$ & \text{$\text{SSIM}$} $\uparrow$ & \text{$\text{RMSE}$} $\downarrow$ & \text{$\text{PSNR}$} $\uparrow$ & \text{$\text{SSIM}$} $\uparrow$ & \text{$\text{RMSE}$} $\downarrow$ \\
     \midrule
     $-$ & \checkmark & $-$ & 37.92 & 0.980 & 4.60 & 39.58 & 0.981 & 2.85 & 34.87 & 0.955 & 3.44 \\
     % \hline
     $-$ & $-$ & \checkmark  & 37.54 & 0.978 & 4.72 & 39.09 & 0.979 & 2.94 & 34.47 & 0.950 & 3.53 \\
     % \hline
     $-$ & \checkmark & \checkmark & 38.17 & 0.981 & 4.48 & 39.82 & 0.982 & 2.82 & 35.13 & 0.956 & 3.37 \\
     % \hline
      \checkmark  & \checkmark  & \checkmark  & 38.52 & 0.981 & 4.37 & 40.03 & 0.982 & 2.76 & 35.42 & 0.958 & 3.30 \\
     \bottomrule
\end{tabular}
}
\end{center}
\end{table*}

\subsection{Complexity Comparison}

Table \ref{tab_Complex} shows the parameter count, computational complexity, inference time, and memory usage. The parameter count (Param(M)) and computational complexity (FLOPs(G)) are calculated using the thop library. Our model ("Ours") has a relatively low parameter count and good FLOPs, striking a balance between computational complexity and performance, with a short inference time.

\begin{table}[t]
\begin{center}
\caption{Complexity comparison with recent models.}
\label{tab_Complex}
\setlength{\tabcolsep}{1mm}
% \fontsize{9}{\baselineskip}\selectfont 
\resizebox{0.49\textwidth}{!}{
      \begin{tabular}{c|c|c|c|c}
        \toprule
        Model & Param(M) & FLOPs(G)	 & Infer(ms) & Mem(GB) \\
        \midrule
         ShadowMaskFormer \cite{li2025shadowmaskformer} & 2.28 & 47.79 & 42 & 9.29    \\
        % \hline
         ShadowDiffusion \cite{guo2023shadowdiffusion} & 55 & 364.12 & 70 & 12.99  \\
        % \hline
         HomoFormer \cite{xiao2024homoformer} & 17.18 & 71.25 & 23 & 7.40  \\
        % \hline
         ShadowFormer-L\cite{guo2023shadowformer} & 11.35 & 129.20 & 38 & 6.93   \\
        \midrule
         % OmniSR & 24.5 & - & -  & - \\
         Ours & 9.39  & 40.09 & 42 & 11.69  \\
       \bottomrule
    \end{tabular}
}

\end{center}
\end{table}

\subsection{Ablation Study}
% For clarity we split the study into two subsections that follow the
% ordering of Tables~\ref{tab_4} and \ref{tab_5}. The first subsection compares alternative scanning strategies; the second removes or replaces individual building blocks to reveal their specific contributions to D2--Mamba.
% \noindent\textbf{Ablation of Scanning Strategies}
% We perform an ablation study to evaluate the effectiveness of various scanning strategies on shadow removal. These strategies differ in how they handle spatial dependencies and contextual integration, each with distinct strengths and weaknesses. The detailed results are shown in Table \ref{tab_4}.
% To comprehensively evaluate the contributions of each proposed submodule within our model, we implemented and tested a series of model variants on the SRD dataset. Table~\ref{tab_4} summarizes the quantitative performance in different configurations. These experiments allow us to gain deeper insight into the specific impact of each component on the final shadow removal performance.

% We conduct ablation studies over SRD to assess the effectiveness of key components in D2-Mamba. As shown in Table~\ref{tab_4}, each proposed component contributes to performance gains.
We conduct ablation studies on SRD to assess the effectiveness of D2-Mamba’s key components. Table~\ref{tab_4} shows that each component contributes to performance gains.
% To thoroughly assess the contribution of each proposed module, we conducted an ablation study over SRD by evaluating several model variants. 
% Results under different settings are summarized in Table~\ref{tab_4}, providing insights into the role of each component in enhancing shadow removal performance.

\noindent\textbf{Effects of DFMB.}
In the encoding stage, we introduced DFMB, which establishes dual-scale aggregation paths to effectively integrate features at different spatial resolutions.
% Specifically, DFMB enhances the model’s ability to capture both global semantics and local details by fusing contextual information from two scales. This mechanism not only enriches the feature representation but also improves the consistency and stability of feature input, thereby providing more discriminative features for the subsequent UNet structure. Ablation results demonstrate that removing DFMB leads to a notable drop in PSNR and an increase in RMSE, particularly affecting the restoration of fine details in complex shadow regions.
By fusing contextual information from both coarse and fine scales, DFMB enhances the capacity to capture global semantics and local details. This not only enriches the feature representation but also stabilizes the input to the UNet, resulting in more discriminative and consistent features. Ablation results show that removing DFMB causes a significant drop in PSNR and a noticeable increase in RMSE, especially in complex regions with fine shadow details.

\begin{table*}[t]
\begin{center}
\caption{Ablation study of the path execution order on the ISTD dataseet.}
\label{tab_5}
\setlength{\tabcolsep}{1mm}
\resizebox{0.8\textwidth}{!}{
    \begin{tabular}{l|ccc|ccc|ccc}
    \toprule
    \multirow{2}{*}{Method} & \multicolumn{3}{c|}{Shadow Region (S)} & \multicolumn{3}{c|}{Non-Shadow Region (NS)} & \multicolumn{3}{c}{All Image (ALL)} \\
                            & PSNR $\uparrow$ & SSIM $\uparrow$ & RMSE $\downarrow$ & PSNR $\uparrow$ & SSIM $\uparrow$ & RMSE $\downarrow$ & PSNR $\uparrow$ & SSIM $\uparrow$ & RMSE $\downarrow$ \\
    \midrule 
          MAS$\rightarrow$HS & 38.98 & 0.991 & 5.75 & 33.86 & 0.980 & 3.92 & 32.04 & 0.968 & 4.20 \\
          HS$\rightarrow$MAS & 39.15 & 0.992 & 5.51 & 33.98 & 0.980 & 3.80 & 32.16 & 0.969 & 4.06  \\
    \bottomrule
    \end{tabular}
}
\vspace{-0.1cm}
\end{center}
\end{table*}

\noindent\textbf{Effects of DPMG.}
% To leverage the structural differences between shadow and non-shadow regions, we designed the DPMG module, combining HS and MAS. Using only HS improves global structural consistency but applies the same processing to both shadow and non-shadow areas, failing to account for the unique properties of the shadow regions, which results in suboptimal performance. On the other hand, MAS focuses on refining the details in the shadow regions, particularly enhancing lighting and color consistency, but lacks the global context needed for effective texture recovery in non-shadow areas, leading to blurry or inconsistent boundaries. By combining HS and MAS, the DPMG module balances both global and local processing: HS strengthens the overall structure, while MAS fine-tunes shadow region recovery. This dual approach significantly improves performance, especially in terms of PSNR and boundary continuity in the shadow regions, while reducing artifacts.
To handle the distinct characteristics of shadow and non-shadow regions, we propose the DPMG, which integrates HS and MAS. Using HS alone promotes global structural consistency but applies uniform processing across the entire image, neglecting the specific needs of shadow regions. Conversely, MAS enhances illumination and color fidelity within shadow areas but lacks sufficient global context, often resulting in texture degradation or boundary blurring. The combined DPMG module leverages the strengths of both: HS preserves structural coherence, while MAS adaptively refines shadow recovery. This design leads to significant gains in PSNR and improved boundary continuity, while effectively reducing visual artifacts.

Building on the design of DPMG, we further investigate how the execution order of the two pathways affects the overall performance. The results of this ablation study, shown in Table~\ref{tab_5}, suggest that the model performs better when the horizontal scanning path precedes the mask-aware adaptive scanning path, rather than the reverse order. This finding reveals an important aspect of how these two pathways interact.
When the horizontal scanning path is executed first, it helps the model capture comprehensive global information, laying a strong foundation for the more detailed and localized shadow recovery in the subsequent mask-aware adaptive scanning stage. With an understanding of the overall image structure, the model can more effectively distinguish between shadowed and non-shadowed regions, improving the accuracy and precision of shadow removal. This global context facilitates more natural integration of the shadow recovery into the image, preserving the overall visual coherence.
On the other hand, if the mask-aware adaptive scanning path is executed first, the model may focus too much on local shadow details while neglecting the broader context. This leads to difficulties in handling complex shadow patterns, resulting in inconsistencies or distortions. The absence of a global structural foundation can cause imprecise shadow removal and disrupt the continuity of the image. Therefore, the ablation results highlight that starting with the horizontal scanning path ensures the model has a better global understanding, which enhances the subsequent mask-aware adaptive scanning and leads to a more accurate and visually coherent shadow removal.

% To better exploit the structural differences between shadow and non-shadow regions, we designed DPMG, which combines the advantages of global horizontal scanning and mask-aware adaptive scanning. At the global level, the horizontal scanning path strengthens contextual relationships between adjacent pixels, aiding in the recovery of texture details and continuity along edges, thereby improving structural clarity. At the local level, the mask-aware scanning path directs the scanning sequence to focus more on intra-class regions (e.g., shadow or non-shadow), enhancing the consistency of lighting and color within shadow areas. Experimental results show that removing DPMG leads to a significant decline in PSNR within shadow regions, and the resulting de-shadowed images are more prone to artifacts such as color inconsistencies or blurred boundaries.

\section{Conclusion}

This paper proposes a novel shadow removal method based on dual-scale fusion and dual-path mask-aware Mamba. The designed DFMB module enhances the model's ability to extract contextual information effectively through a dual-scale fusion mechanism, stabilizing the diversity of input features. The DPMG module combines the advantages of global horizontal scanning and shadow-priority scanning, effectively leveraging global information while ensuring local consistency, thereby significantly improving the recovery of shadow regions. Extensive experimental evaluations on SRD, ISTD, and ISTD+, demonstrate that our method achieves leading performance across various evaluation metrics, particularly excelling in detail preservation and restoration of image brightness and color consistency. These results indicate that our method provides an efficient and robust solution for the shadow removal task. Future research could further explore the potential in more complex scenarios and shadow patterns, thereby improving the generalizability.

\bibliographystyle{plain} 
\bibliography{reference}

\end{document}